%% file: main.tex
\documentclass[letterpaper, 10 pt, conference]{ieeeconf}  % Comment this line out if you need a4paper

\IEEEoverridecommandlockouts                              % This command is only needed if 
\usepackage{graphics} % 图像
\usepackage{amsmath} % 数学公式
\usepackage{amssymb} % 数学符号
\usepackage{pgfplots}  
\usepackage{multirow}
\usepackage{adjustbox}
\usepackage{booktabs}
\usepackage{hyperref}
\usepackage{url}
\pgfplotsset{compat=1.18}
\title{\LARGE \bf 
ERetinex: Event Camera Meets Retinex Theory for Low-Light Image Enhancement
}
\author{
    Xuejian Guo,
    Zhiqiang Tian, Member, IEEE,
    Yuehang Wang,
    Siqi Li,
    Yu Jiang, Member, IEEE,\\
    Shaoyi Du, Member, IEEE,
    and Yue Gao*, Senior Member, IEEE.\\
\thanks{Xuejian Guo and Zhiqiang Tian are with the School of Software
Engineering, Xi’an Jiaotong University, Xi’an 710049, China (e-mail:
lodew920@stu.xjtu.edu.cn; zhiqiangtian@xjtu.edu.cn).}
\thanks{Yu Jiang and Yuehang Wang are with the College of Computer Science and Technology, Jilin University, Changchun 130012, China (e-mail:jiangyu2011@jlu.edu.cn; yuehang22@mails.jlu.edu.cn).}
\thanks{Shaoyi Du is with the National Key Laboratory of Human-Machine
Hybrid Augmented Intelligence, National Engineering Research Center
for Visual Information and Applications, and Institute of Artificial Intelligence and Robotics, Xi’an Jiaotong University, Xi’an 710049, China (e-mail: dushaoyi@xjtu.edu.cn).}
\thanks{Siqi Li and Yue Gao are with BNRist, THUIBCS, KLISS, BLBCI, School of Software, Tsinghua University, Beijing 100084, China (lsq19@mails.tsinghua.edu.cn; kevin.gaoy@gmail.com).}
\thanks{*Corresponding author: Yue Gao}
}
\begin{document}
\maketitle
\thispagestyle{empty}
\pagestyle{empty}
% ABSTRACT
\input{sec/0_abstract}
% INTRODUCTION
\input{sec/1_intro}
% RELATE
\input{sec/2_relate}
% METHOD
\input{sec/3_method}
% EXPERIMENT
\input{sec/4_experiments}
% CONCLUSIONS
\input{sec/5_conclusion}
% \addtolength{\textheight}{-12cm}
% \section*{APPENDIX}
% \section*{ACKNOWLEDGMENT}
\bibliographystyle{IEEEtran}
\bibliography{IEEEabrv}
\end{document}

%% file: sec/0_abstract.tex
\begin{abstract}
Low-light image enhancement aims to restore the under-exposure image captured in dark scenarios. Under such scenarios, traditional frame-based cameras may fail to capture the structure and color information due to the exposure time limitation. Event cameras are bio-inspired vision sensors that respond to pixel-wise brightness changes asynchronously. Event cameras' high dynamic range is pivotal for visual perception in extreme low-light scenarios, surpassing traditional cameras and enabling applications in challenging dark environments. In this paper, inspired by the success of the retinex theory for traditional frame-based low-light image restoration, we introduce the first methods that combine the retinex theory with event cameras and propose a novel retinex-based low-light image restoration framework named ERetinex. Among our contributions, the first is developing a new approach that leverages the high temporal resolution data from event cameras with traditional image information to estimate scene illumination accurately. This method outperforms traditional image-only techniques, especially in low-light environments, by providing more precise lighting information. Additionally, we propose an effective fusion strategy that combines the high dynamic range data from event cameras with the color information of traditional images to enhance image quality. Through this fusion, we can generate clearer and more detail-rich images, maintaining the integrity of visual information even under extreme lighting conditions. The experimental results indicate that our proposed method outperforms state-of-the-art (SOTA) methods, achieving a gain of 1.0613 dB in PSNR while reducing FLOPS by \textbf{84.28}\%. The code is available at \textcolor{red}{\url{https://github.com/lodew920/ERetinex}}.
\end{abstract}

%% file: sec/1_intro.tex
\section{Introduction}
In dark scenarios, reconstructing clear images is an important and challenging task in computer vision. Low-light image enhancement is crucial for robotics, particularly for applications in autonomous navigation, night-time surveillance, and precise 3D environment mapping. It significantly improves visibility and accuracy in dark conditions, empowering robots to perform various tasks with reliability and efficiency. Many algorithms for enhancing low-light images have been proposed, gradually evolving from traditional simple methods such as gamma correction and histogram equalization to methods based on retinex theory. 
\input{figure/efficient}
Unlike directly using neural networks to learn a brute-force mapping function from low-light images to normal-light images, retinex-based neural networks can better satisfy human colour perception. Retinex theory \cite{land1971lightness} assumes that a colour image can be decomposed into reflectance and illumination components, treating enhancement as a parameter estimation problem. With the advancement of deep learning, convolutional neural networks have been applied to low-light image enhancement \cite{Zhu_Zhang_Shen_Ma_Zhao_Zhou_2020, Wang_Zhang_Fu_Shen_Zheng_Jia_2019, Wei_Wang_Yang_Liu_2018} using various strategies to decompose colour images, denoise reflectance, and adjust the illumination. The recent rise of deep learning models like Transformers \cite{Cai_2023_ICCV, Wu_2023_CVPR, Fu_2023_CVPR} offers a potential solution to the shortcomings of CNN-based methods. However, the limitations of visual information data captured by frame-based cameras in such low-light conditions can cause detailed information to be lost, thus limiting image restoration.

Unlike traditional cameras, event cameras are biologically inspired visual sensors that asynchronously trigger events based on intensity changes, outputting event streams to record scene information. It can reflect changes in illumination and can also depict structural information, which is important for low-light image enhancement. They offer advantages such as high dynamic range, high temporal resolution, no motion blur, and so on. Therefore, event cameras are widely used in various computer tasks, such as image restoration \cite{Jiang_Wang_Li_Zhang_Zhao_Gao}, video super-resolution \cite {Lu_2023_CVPR}, optical flow \cite{Luo_2023_ICCV}, low-power tracking \cite{Zhu_2023_ICCV}, deblurring \cite{Sun_2023_CVPR}, \textit{etc.} In the field of image enhancement, Wang \textit{et al.} \cite{Jiang_Wang_Li_Zhang_Zhao_Gao} proposed an event-based RGB image enhancement framework ELLIM and built the image-event paired dataset. However, the framework is redundant and does not utilize events well.

To cope with the above problems, we introduce a novel approach that combines event cameras with the retinex theory for low-light image enhancement, introducing a lightweight framework named ERetinex. Specifically, Our ERetinex estimates lighting information, captures lost reflection information, and uses it to illuminate and restore low-light images. In addition, we present an efficient fusion technique that integrates event camera data with traditional images. It first recovers target images using colour information from the images and then enhances low-light images using structural information from the event camera, improving clarity and detail while maintaining visual integrity in low-light conditions. As shown in Fig. \ref{Fig:1}, our lightweight ERetinex outperforms SOTA image enhancement methods with fewer parameters. 

Our contributions are summarized as follows:
 
\begin{itemize}
\item For the first time, we combine events with retinex theory and design a lightweight framework, called ERetinex that achieves SOTA performance while reducing FLOPS by 84.28
\%.
\item We develop a new method that uses high temporal resolution data from event cameras together with traditional image data to estimate scene illumination accurately.
\item  We propose a strategy to fuse the structural information of event cameras with the colour information of traditional images. This fusion improves image quality and maintains visual information integrity under extreme lighting conditions.
\end{itemize}

%% file: figure/efficient.tex
\begin{figure}[t]
  \centering
  \includegraphics[width=0.9\linewidth, trim=2 2 2 2, clip]{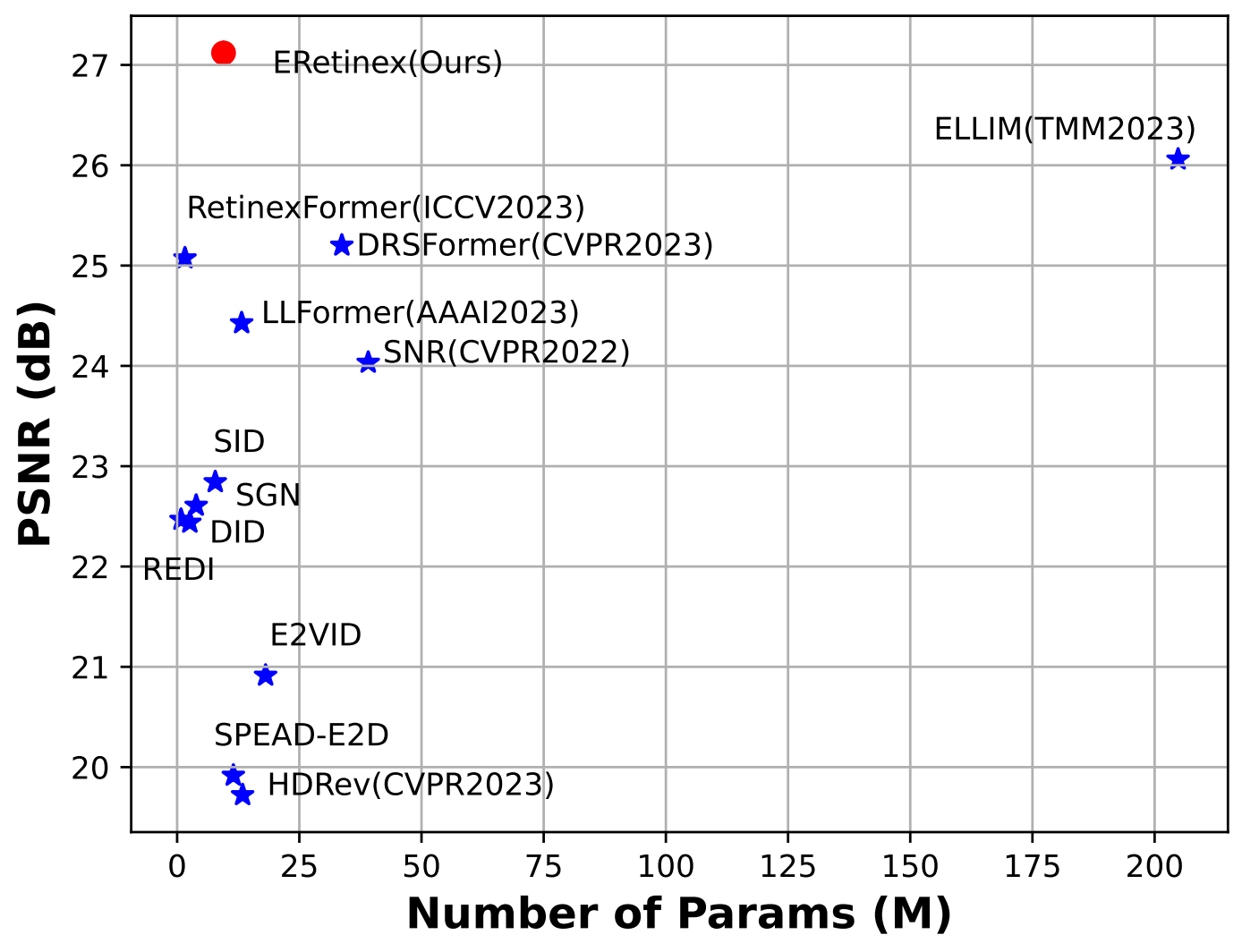}
  \caption{Our ERetinex attains SOTA performance in low-light image reconstruction, demonstrating superior results with fewer parameters.}
  \label{Fig:1}
\end{figure}

%% file: sec/2_relate.tex
\section{Related Work}
\subsection{Retinex-based Low-light Image Enhancement} 
The retinex theory was proposed a long time ago, based on the color invariance model and the subjective perception of color by the human visual system \cite{land1971lightness}.
It decomposes the image into brightness and reflectance components and then processes each component independently. By separating brightness and reflectance information, this approach can significantly improve image quality, making details more clearly visible.

\textbf{Traditional Methods.} In the past, retinex-based methods emerged in endlessly \cite{Guo_Li_Ling_2017, land1971lightness, Lee_Shih_Lien_Han_2013, Ren_Li_Cheng_Liu_2018, Wang_Zheng_Hu_Li_2013}. For instance, LIME \cite{Guo_Li_Ling_2017} estimates the illumination map by initializing it with the maximum values across the three color channels and then refining it using a structural prior to obtain the final lighting map. MSR \cite{Jobson_Rahman_Woodell_1997} designed a Gaussian filter or a set of filter banks to achieve resolution of illumination and reflection patterns. JED \cite{Ren_Li_Cheng_Liu_2018} enhances the image and suppresses noise by combining sequence decomposition and gamma transform. To address degradation caused by noise, Li \textit{et al.} \cite{Li_Liu_Yang_Sun_Guo_2018} incorporated a noise term into the objective function, enabling simultaneous noise removal and detail enhancement.
% However, conventional model-based methods most rely on carefully designed hand-craft priors or certain statistical models, they are not robust against degraded visibility and unexpected noise. Besides, these methods naively assume that the low-light images are corruption-free, leading to severe noise and color distortion issues during the enhancement process.

\textbf{Modern deep learning methods.} Deep learning introduces efficient solutions to traditional low-light enhancement, employing end-to-end networks for direct enhancement and deep Retinex-based methods to automatically extract features for image improvement. KinD \cite{Zhang_Zhang_Guo_2019} employs a decomposition and adjustment framework, utilizing convolutional neural networks to learn the mappings for both stages. Similarly, Retinex-Net \cite{Wei_Wang_Yang_Liu_2018} integrates the Retinex decomposition paradigm with deep learning techniques. Wang \textit{et al.} \cite{Wang_Zhang_Fu_Shen_Zheng_Jia_2019} propose a one-stage retinex-based CNN, dubbed DeepUPE, to directly predict the illumination map. However,  these methods cannot make full use of global information. In addition, Cai \textit{et al.} \cite{Cai_2023_ICCV} propose a lighting-guided transformer to guide lighting recovery and resolve noise and impurities. Yi \textit{et al.} \cite{Yi_2023_ICCV} designed a transformer decomposition network to obtain the reflection map. Fu \textit{et al.} \cite{Fu_2023_CVPR} use transformer in lighting module and compositing module. Wu \textit{et al.} \cite{Wu_2023_CVPR}  design a semantic-aware transformer for low-light image enhancement. These methods effectively mitigate corruptions like noise, artifacts, and color distortion, whether inherent in low-light conditions or introduced during enhancement, yet some information loss is unavoidable.
\subsection{Event-based image restoration} 
Event cameras have the characteristics of high dynamic range, high temporal resolution, and low latency, they are widely used in various computer vision tasks, such as image restoration \cite{Jiang_Wang_Li_Zhang_Zhao_Gao}, Optical Flow \cite{Luo_2023_ICCV}, Tracking \cite{Zhu_2023_ICCV}, Motion Estimation \cite{Huang_2023_CVPR}. For example, Zhang \textit{et al.} \cite{zhang2020learning} investigated the capabilities of event cameras in low-illumination environments, employing an unsupervised domain adaptation strategy to achieve grayscale image restoration. Meanwhile, Zhou \textit{et al.} \cite{Zhou_Teng_Han_Xu_Shi_2021} introduced a two-stage approach that leverages light streaks and local events to address the issue of image blurring. In the field of event-based RGB image restoration, \cite{Jiang_Wang_Li_Zhang_Zhao_Gao}, \cite{liu2023low}, \cite{liang2023coherent} use event and low-light images to restore the original RGB image.

%% file: sec/3_method.tex
\section{Method}
Fig. \ref{fig:2} illustrates the overall architecture of our method. ERetinex utilizes two distinct inputs: low-light image and event. The low-light image undergoes a feature extraction process to generate image features, while the event data is voxelized and subjected to feature extraction. Subsequently, the image features and event features are combined, creating a fused feature set employed to estimate illumination information. Next, the illuminated image will be fed into the event and image fusion module. The low-light image will be initially enhanced using the color information in the image, and then further refined and improved in detail using the structural information in the event data
\input{figure/pipeline}
\subsection{Event representation}
The event camera fires events asynchronously at each pixel, where each event is represented as a quad $e_i = [x_i,y_i,t_i,p_i]$, where $x_i$  and $y_i$ represent the pixel coordinates, $t_i$ represents the timestamp of the event. Similar to \cite{rebecq2019high, Jiang_Wang_Li_Zhang_Zhao_Gao}, we encode the event stream to the 3D voxel grid. Specifically, given an event stream $\psi = \{e_i\}_{i=1}^{N}$, where N means the the number of the events. We adopt a non-overlapping time window $\delta = t_N-t_1$ to divide $\psi$ into B temporal bins, and then allocate the polarity of each stream to the two closest spatiotemporal neighbors as follows:
\begin{equation}
E_{(x_l,y_m,t_n)}=\sum_{\substack{x_i=x_l \\ y_i=y_m}} p_i \cdot \text{max}(0, 1-\lvert t_n -t_i^\ast\rvert),
\label{eq:1}
\tag{1}
\end{equation}
where $t_i^\ast = \frac{t_i - t_{1}}{t_{N}-t_{1}} \times (B-1) $. Given that the event's head and tail contain significant noise but limited information, we have set $B$=7. We discard the first and last bins, retaining only the middle five bins as the event data. Eventually, we convert the event stream to a grid-based tensor $E \in \mathbb{R}^{5 \times W \times H}$, where $W$ and $H$ are the spatial sizes of the low light image.
\subsection{Retinex-based Architecture}
According to the retinex theory, an image $I\in \mathbb{R}^{H\times W\times3}$ can be decomposed into a reflectance image $R \in \mathbb{R}^{H\times W\times3}$ and an illumination map $L\in \mathbb{R}^{H\times W}$ as 
\begin{equation}
I = R \odot L,
\label{eq:2}
\tag{2}  
\end{equation}
\input{table/comparison_method}
where $\odot$ means the element-wise multiplication. However, considering $I$ as entirely corruption-free is inconsistent with real-world scenarios. In practical low-light environments, the inherent limitations of equipment configurations and settings used for capturing scenes, including high ISO settings and extended exposure times, inevitably lead to the presence of noise, undesirable artefacts, and colour distortion. To more accurately model these noises, we introduce perturbation terms for $R$ and $L$ in Eq. \eqref{eq:2}, as 
\begin{equation}
\begin{aligned}
I &= (R + \overline{R})\odot  (L + \overline{L}) \\
  &= R \odot L + R \odot \overline{L} + \overline{R} \odot (L + \overline{L}),
\end{aligned}
\label{eq:3}
\tag{3}
\end{equation}
\input{figure/image_event_fuse}
where $\overline{R}$ is the same shape as $R\in \mathbb{R} ^ { H \times W \times 3}$, and $\overline{L}$ is the same shape as $L\in \mathbb{R} ^ {H \times W}$. Similar to \cite{Fu_Zeng_Huang_Zhang_Ding_2016, Guo_Li_Ling_2017,  Wang_Zhang_Fu_Shen_Zheng_Jia_2019}, we regard $R$ as a well-exposed image. To light up $I$, we element-wisely multiply the two sides of Eq. \eqref{eq:3} by a light-up map $L^{-1} \in \mathbb{R} ^ {H \times W}$ such that  $L\odot L^{-1}=1$ as 
\begin{equation}
\overline{I} = I \odot L^{-1} = R + C,
\label{eq:4}
\tag{4}
\end{equation}
where $C = R \odot(\overline{L} \odot L^{-1}) + (\overline{R} \odot (L + \overline{L}))\odot L^{-1}$. Similar to \cite{Cai_2023_ICCV}, $C$ means these corruptions are amplified by ${L\textsuperscript{-1}}$. 
Subsequently, we formulate our ERetinex as
\begin{equation}
\begin{aligned}
L^{-1}  &= \varepsilon(Image, Event) \\
I_{mid} &= \eta(I \odot L^{-1}, Image), \\
I_{out} &= \Phi(I_{mid}, Event)
\end{aligned}
\label{eq:5}
\tag{5}
\end{equation}
where $\varepsilon$ is an illumination estimator, it takes $Image$ and $Event$ (as shown in \eqref{eq:1}) as inputs and outputs a lighting map ($L^{-1}$). Then $L^{-1}$ is fed into $\eta$ to restore the corruptions like colour with image, and produce the initial enhanced image $I_{mid} \in \mathbb{R}^{H\times W\times 3}$. After that, $I_{mid}$ is fed into $\Phi$ to restore the corruptions like edges and textures with events and produce the enhanced image $I_{out} \in \mathbb{R}^{H\times W\times 3}$.

\subsection{Illumination Estimator}
In our research, we introduce a new approach to scene illumination estimation (as shown in Fig. \ref{fig:2} (i)) that integrates the structure data of event cameras with the rich color information captured by traditional frame-based cameras. This synergistic combination is pivotal for enhancing the accuracy of illumination estimation, particularly in challenging low-light conditions where conventional methods often fall short. Our method capitalizes on the unique properties of event cameras, which are capable of continuously monitoring changes in the scene's luminance and generating events asynchronously. Each event records a timestamp and the pixel location where a significant change in brightness occurs, allowing for the capture of motion and changes in lighting with unprecedented precision. This event-driven data acquisition is complemented by the colour and intensity information provided by traditional cameras, which offer a complete snapshot of the scene at regular intervals.
\subsection{Dual-Modality Enhancement Strategy}
In this study, we design an efficient fusion strategy that fully leverages the complementary advantages of event cameras and conventional image sensors. This strategy combines the high dynamic range and high temporal resolution characteristics of event cameras with the rich color information captured by traditional image sensors. As illustrated in Fig. \ref{fig:2} (ii), the fusion process is divided into two steps: first, using the color data of traditional images to preliminarily enhance the brightness of low light scenes; Then, the structural information captured by the event camera is used to further refine the image, enhance details, and reveal subtle features that may be lost in complex lighting environments. The core of this two-stage enhancement method lies in utilizing the high temporal resolution of event cameras, which can quickly respond to scene changes and provide a solid foundation for dynamic range expansion. At the same time, we maintain the color accuracy of traditional images, ensuring that the final image meets high standards in both visual appeal and semantic accuracy. Fig. \ref{fig:3} provides a detailed schematic diagram of a low light image restoration framework guided by image and event data. Initially, features are downsampled to reduce spatial dimensions, and then deep convolutions are performed to independently extract local spatial features for each channel. Subsequently, the features are upsampled to restore the original spatial resolution, during which the channel attention mechanism is applied to recalibrate the feature map, emphasizing prominent features and suppressing noise. In each BLOCK, there will be a guiding feature for an event or image and an output feature from the previous layer. The output feature from the previous layer will undergo deep analyzable convolution and channel level attention, and the image and event will be guided in the attention.

%% file: figure/pipeline.tex
\begin{figure*}[thpb]
    \centering
    \includegraphics[width=0.9\linewidth, trim=30 333 495 25, clip]{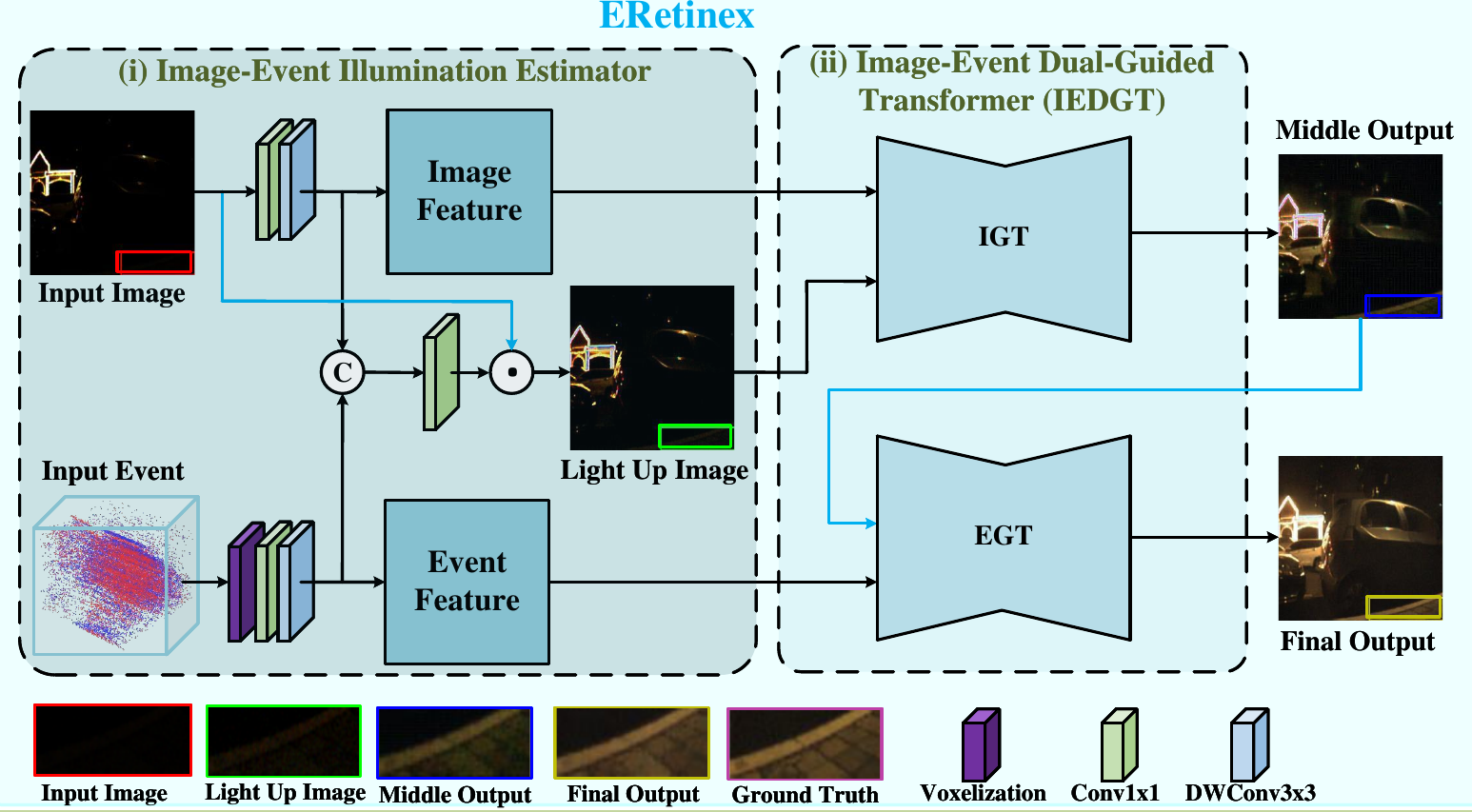}
    \caption{The overview of our method. ERetinex illustrates the framework of integrating event-based retinex theory, which consists of an image-event illumination estimator (i) and an Image-Event Dual-Guided Transformer (IEDGT) (ii)}
    \label{fig:2}
\end{figure*}

%% file: table/comparison_method.tex
\begin{table*}
\renewcommand{\arraystretch}{1.2}
\caption{Quantitative Comparisons on LIE (indoor, outdoor, and total) datasets. The highest result is in \textcolor{red}{red} colour, while the second highest result is in \textcolor{blue}{blue} colour. Our ERetinex outperforms SOTA algorithms.}
\resizebox{\textwidth}{!}{
\begin{tabular}{ l | c c | c c c | c c c | c c c }
\toprule
\multirow{2}{*}{\textbf{Methods}} & 
\multicolumn{2}{c|}{\textbf{Complexity}} & 
\multicolumn{3}{c|}{\textbf{Indoor}} &
\multicolumn{3}{c|}{\textbf{Outdoor}} & 
\multicolumn{3}{c}{\textbf{Total}}
\\
\cline{2-12}
&\textbf{FLOPS (G)}&\textbf{Param (M)}&\textbf{PSNR$\uparrow$} & \textbf{SSIM$\uparrow$} & \textbf{LPIPS$\downarrow$} & \textbf{PSNR$\uparrow$} & \textbf{SSIM$\uparrow$} & \textbf{LPIPS$\downarrow$} & \textbf{PSNR$\uparrow$} & \textbf{SSIM$\uparrow$} & \textbf{LPIPS$\downarrow$}
\\
\midrule
E2VID\cite{rebecq2019high} & 18.15 & 18.1 & 22.0002 & 0.7661 & 0.4276 & 16.6336 & 0.6750 & 0.4919 & 20.9133 & 0.7477 & 0.4406 \\
SPEAD-E2VID\cite{cadena2021spade}  & 96.22 & 11.5 & 20.9368 & 0.7612 & 0.4305 & 15.8976 & 0.6908 & 0.4920 & 19.9162 & 0.7470 & 0.4429 \\
\hline
DID\cite{maharjan2019improving} & 38.95 & 2.6 & 23.0800 & 0.8019 & 0.4329 & 19.9129 & 0.8020 & 0.4233 & 22.4386 & 0.8019 & 0.4309 \\
SID\cite{Chen_2018_CVPR} & 2.82 & 7.8 & 23.4609 & 0.8081 & 0.4401 & 20.4065 & 0.8086 & 0.4329 & 22.8422 & 0.8082 & 0.4387 \\
SGN\cite{gu2019self} & 12.95 & 3.9 & 23.2162 & 0.7911 & 0.4519 & 20.2109 & 0.8099 & 0.4087 & 22.6076 & 0.7949 & 0.4432 \\
REDI\cite{lamba2021restoring} & 0.300 & 0.8 & 23.3305 & 0.7933 & 0.4705 & 19.0679 & 0.7381 & 0.5123 & 22.4672 & 0.7821 & 0.4789 \\
SNR\cite{xu2022snr} & 22.69 & 39.1 & 24.3627 & 0.8201 & 0.4253 & 22.7356 & 0.8491 & 0.3594 & 24.0331 & 0.8259 & 0.4120 \\
DRSFormer\cite{Chen_2023_CVPR} & 226.30 & 33.7 & 25.3917 & 0.8468 & 0.3686 &  \textcolor{blue}{24.4496} & 0.8759 & 0.3127 & 25.2009 & 0.8527 & 0.3573 \\
LLFormer\cite{wang2023ultra} & 23.84 & 13.2 & 24.8294 & 0.8334 & 0.3946 & 22.8411 & 0.8468 & 0.3772 & 24.4267 & 0.8361 & 0.3911 \\
RetinexFormer\cite{Cai_2023_ICCV} & 15.57 & 1.6 & 25.6733 & 0.8490 & 0.3835 & 22.7111 & 0.8403 & 0.3397 & 25.0734 & 0.8472 & 0.3747 \\
\hline
ELLIM\cite{Jiang_Wang_Li_Zhang_Zhao_Gao} & 217.81 &204.8 & \textcolor{blue}{26.5926} & \textcolor{blue}{0.8792} & \textcolor{blue}{0.2690} & 24.0294 & \textcolor{blue}{0.8774} & \textcolor{blue}{0.2774} & \textcolor{blue}{26.0586} & \textcolor{blue}{0.8788} & \textcolor{blue}{0.2707} \\
Ours & 34.23 & 9.5 & \textcolor{red}{27.6071} & \textcolor{red}{0.8823} &  \textcolor{red}{0.2406} & \textcolor{red}{25.2013} & \textcolor{red}{0.8921} & \textcolor{red}{0.2541} & \textcolor{red}{27.1199} & \textcolor{red}{0.8843} & \textcolor{red}{0.2433} \\
\bottomrule
\end{tabular}
}
\label{tab:1}
\end{table*}

%% file: figure/image_event_fuse.tex
\begin{figure}[thpb]
    \centering
    \includegraphics[width=0.99\linewidth, trim=16 60 330 10, clip]{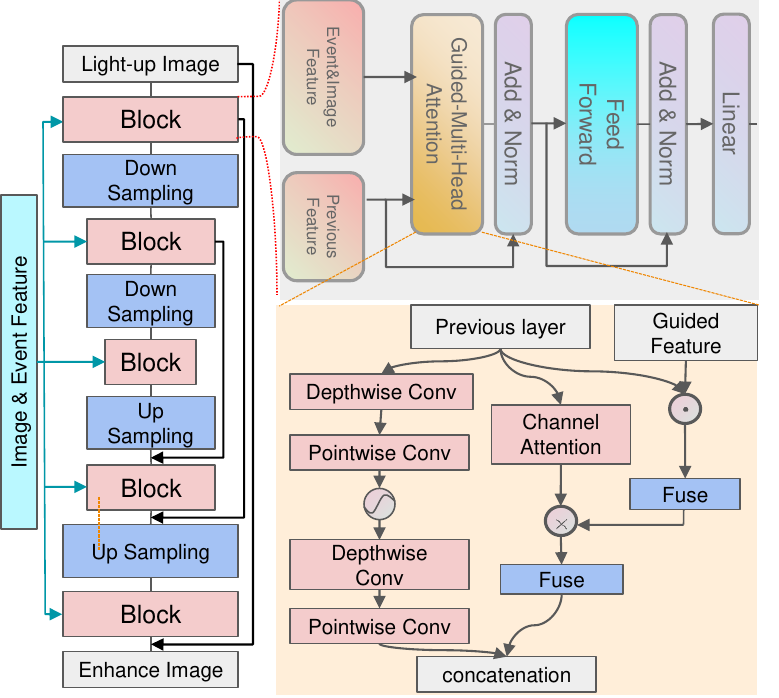}
    \caption{Illustration of Low-Light Image Restoration Guided by Image and Event Data.}
    \label{fig:3}
\end{figure}

%% file: sec/4_experiments.tex
\section{Experiment}
\input{figure/comparev1}
\subsection{Datasets and Implementation Details}
% \textbf{Implementation Details.} Our method is implemented in PyTorch on an NVIDIA 3090 GPU. Training uses the Adam optimizer \cite{Kingma_Ba_2014} with an initial learning rate of $1\times10^{-4}$
% , reduced to $1\times10^{-6}$ via cosine annealing \cite{Loshchilov_Hutter_2016}. We train on the full LIE dataset and test on indoor, outdoor, and combined scenes. Images are cropped to $256 \times 256$ with a batch size of 2. The loss combines mean absolute error (MAE) and Learned Perceptual Image Patch Similarity (LPIPS) between enhanced and ground truth images. Evaluation metrics include PSNR, SSIM, and LPIPS.
\textbf{Implementation Details.} Our proposed method is implemented in PyTorch and executes on the NVIDIA 3090 GPU. The Adam \cite{Kingma_Ba_2014} optimizer is utilized during training with an initial learning rate of $1\times10^{-4}$, which is adaptively reduced to $1\times 10^{-6}$ by cosine annealing scheme \cite{Loshchilov_Hutter_2016}. We train on the full LIE dataset and test on indoor, outdoor, and combined scenes. During training and testing, the crop size of all images is adjusted to $256 \times 256$, and the batch size is set to 2. The training objective is to minimize the mean absolute error (MAE) and Learned Perceptual Image Patch Similarity (LPIPS) between enhanced image and ground truth. We adopt the peak signal-to-noise ratio (PSNR), structural similarity (SSIM), and LPIPS as the evaluation metrics.

\textbf{Dataset.} We selected the challenging LIE dataset \cite{Jiang_Wang_Li_Zhang_Zhao_Gao}, a real-world dataset that pairs event and low-light images, with the majority of captured scenes having a luminosity falling below 2 lux. The LIE benchmark comprises 2218 short-exposure RGB images and their corresponding event pairs. Of these, 1981 pairs are designated for training, while 237 pairs are allocated for testing. Within this dataset, the indoor dataset encompasses 1634 pairs, while the remaining 189 pairs are reserved for testing. The outdoor dataset includes 347 pairs for training and 48 pairs for testing. 
\subsection{Low-light Image Enhancement}
\textbf{Quantitative Results.} Our method is benchmarked against leading SOTA enhancement algorithms, spanning low-light image improvement (DID, SID, LLFormer, RetinexFormer), event-based techniques (E2VID, SPEAD-E2VID), combined event and low-light approaches (ELLIM), and general low-level task algorithms (DRSFormer). The PSNR, SSIM, and LPIPS results are reported in Tab. \ref{tab:1}. ERetinex achieves superior performance over SOTA methods while maintaining low computational and memory requirements. Compared to the recent SOTA method ELLIM \cite{Jiang_Wang_Li_Zhang_Zhao_Gao}, which leverages both low-light images and events, our ERetinex attains improvements of 1.0145 dB, 1.1719 dB, and 1.0613 dB in indoor, outdoor, and total scenarios, respectively. Remarkably, ERetinex achieves these gains while only requiring a modest \textbf{4.64\%} (9.5/204.8) of the parameters and \textbf{15.72\%} (34.23/217.81) of the FLOPS in comparison.

\textbf{Qualitative Results.} The visual results of testing ERetinex and SOTA algorithms are presented in Fig. \ref{fig:4}. Please zoom in for a closer examination. Given that events lack color information, employing events directly to restore colour images often leads to colour distortions, as seen in E2VID. The most advanced image restoration techniques available today, such as LLFormer, DRSFormer, RetinexFormer and ELLIM fall short in accurately preserving fine details, as demonstrated in the magnified illustration. In contrast, ERetinex enhances visibility in low-contrast conditions, eliminating noise while preserving color and fine details.

\subsection{Ablation Study}
We conduct an ablation study on the LIE Total dataset to assess the roles of images, events, guidance modules, fusion strategies, and key components like channel attention and depth convolutions in ERetinex.
\input{table/abla_light}

\textbf{Illumination Estimator.} To investigate the roles of events and images in illumination estimation, we perform ablation experiments. As shown in Tab. \ref{Tab:2}, removing the image reduces the PSNR by 0.8286, while removing the event decreases it by 0.8094. When both are excluded, the PSNR drops significantly by approximately 1.2535. These findings demonstrate that both images and events are crucial for accurate illumination estimation, highlighting the limitations of relying solely on images for lighting estimation.
\input{table/abla_fuse_order}
\input{figure/fuse}

\textbf{The sequence of guiding images and events.}
We give priority to using the image guide first, followed by the event guide (as illustrated in Fig. \ref{fig:5} (Series I)). To explore different fusion mechanisms, we conduct ablations on the fusion mechanism, as shown in Fig. \ref{fig:5} (Series II and Parallel), and the corresponding results are presented in Tab. \ref{Tab:3}. From the results, it is evident that during the low-light enhancement process, events and images play highly complementary roles. In addition to capturing the contours and edges of objects, event data also contains various types of noise, which can be detrimental to image recovery. Therefore, if event data is first used as the main input for image restoration, it may lead to learning impurities in the output image, thus misleading subsequent feature learning. When we allow IGT and EGT to extract features in parallel, the PSNR decreases. This is due to insufficient noise-handling capabilities of images and events when processed in parallel. Since images contain rich colour information, it is preferable to initially employ images for guiding the recovery of texture and colour. Subsequently, events are utilized to further restore the structure and contours of the image. 
\input{table/abla_role}

\textbf{Role of Image and Event.} Tab. \ref{Tab.4} summarizes an ablation study assessing the contributions of image data, event data, and key model components to low-light image enhancement. The baseline model achieves a PSNR of 27.1199 and an SSIM of 0.8843. Removing image data (W/o Image) causes a significant performance drop, reducing PSNR to 23.2375 and SSIM to 0.8511, emphasizing the critical role of image information. Excluding event data (W/o Event) results in a moderate decline, with PSNR at 25.4549 and SSIM at 0.8580, indicating its secondary yet valuable contribution. Without image guidance (W/o Image Guide), PSNR drops to 26.7900 and SSIM remains at 0.8844, while removing event guidance (W/o Event Guide) reduces PSNR to 25.7602 and SSIM to 0.8697, showing that both guidance mechanisms enhance performance, with image guidance having a stronger impact. Removing channel attention (W/o Channel Attention) drastically reduces performance (PSNR: 23.4301, SSIM: 0.8260), while replacing depthwise separable convolution (W/o DwConv) slightly lowers results (PSNR: 26.0102, SSIM: 0.8718), emphasizing the importance of these components in low-light enhancement.

%% file: figure/comparev1.tex
\begin{figure*}
    \centering
    \includegraphics[width=0.9\textwidth, trim=5 5 5 5, clip]{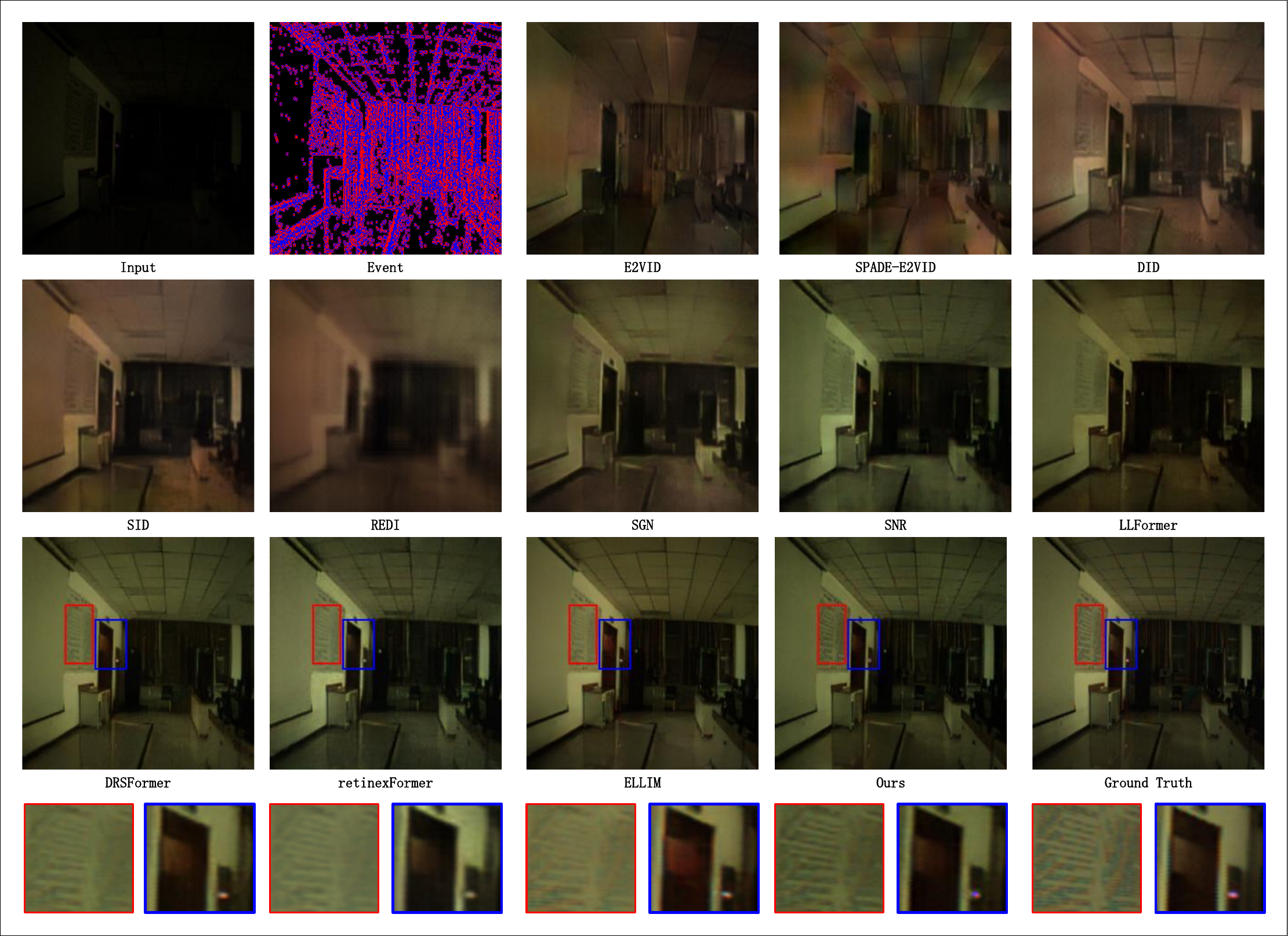}
    \caption{In this indoor scene, multiple algorithms demonstrate their recovery results. Models with superior performance are enlarged for comparison. It is evident that our model consistently outperforms the others in detail and colour reproduction.}
    \label{fig:4}
\end{figure*}

%% file: table/abla_light.tex
\begin{table}[h]
\caption{Ablation studies analyze the roles of image and event data in illumination estimation.}
\vspace{-1em}
\label{Tab:2}
\begin{center}
\begin{tabular}{lccccc}
    \toprule
    \textbf{} & \textbf{PSNR} & \textbf{SSIM} \\
    \midrule
    Baseline & \textbf{27.1199} & \textbf{0.8843} \\
    W/o Image illumination & 26.2913 & 0.8737\\
    W/o Event illumination & 26.3105 & 0.8764  \\
    W/o Both & 25.8664 & 0.8743 \\
    \bottomrule
  \end{tabular}
\end{center}
\vspace{-1em}
\end{table}

%% file: table/abla_fuse_order.tex
\begin{table}[h]
\caption{The result of three different fusing methods in Fig. \ref{fig:5}.}
\label{Tab:3}
\vspace{-1em}
\begin{center}
\begin{tabular}{lccc}
    \toprule
    \textbf{} & \textbf{Indoor} & \textbf{Outdoor} & \textbf{Total}\\
    \midrule
    Series I  & \textbf{27.6071} & \textbf{25.2013} & \textbf{27.1199}\\
    Series II & 26.9256        & 24.6225          & 26.4594\\
    Parallel  & 25.7695        & 23.5614          & 25.3223\\
    \bottomrule
\end{tabular}
\end{center}
\vspace{-1em}
\end{table}

%% file: figure/fuse.tex
\begin{figure}
    \centering
    \includegraphics[width=0.9\linewidth, trim=6 6 6 6, clip]{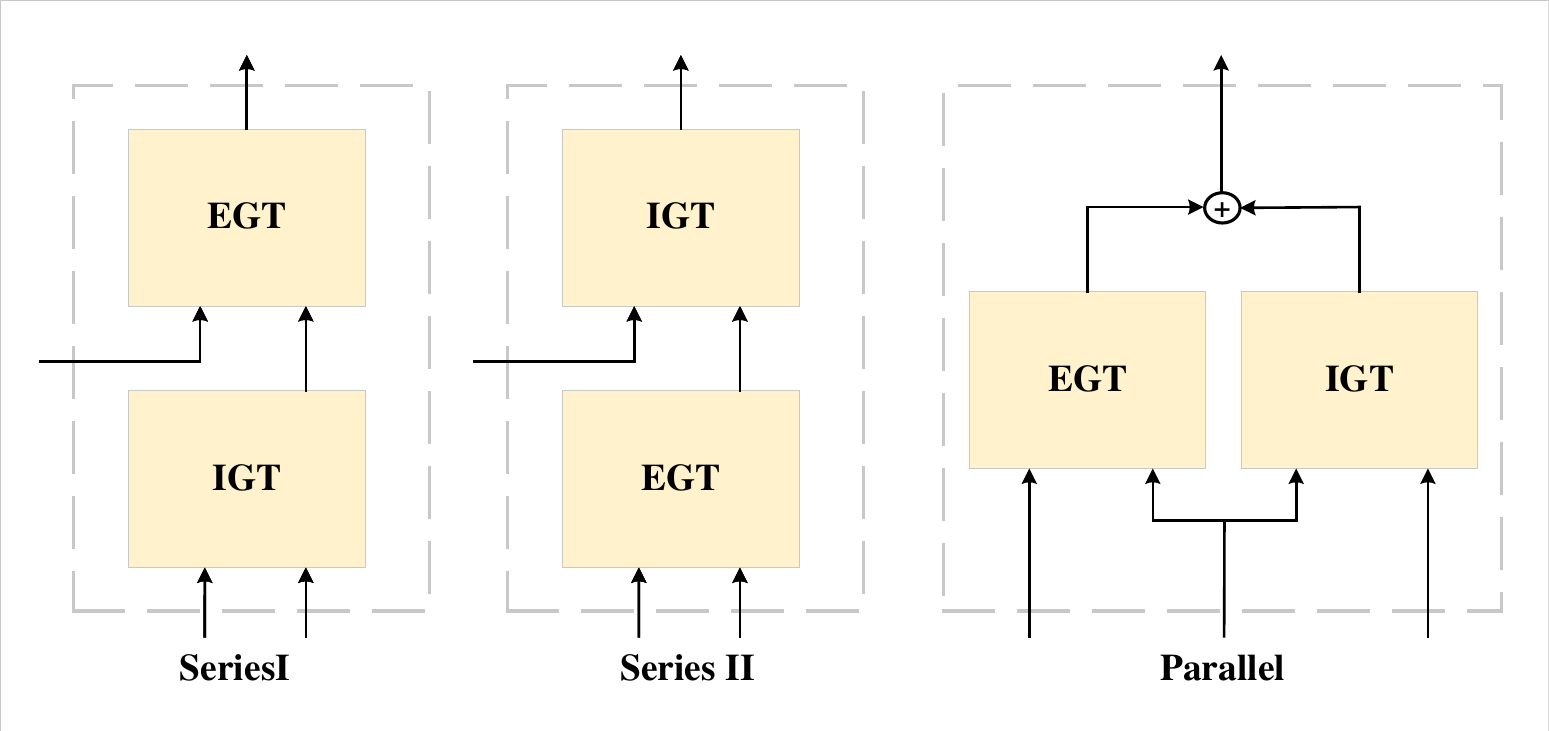}
    \caption{Three approaches to integrate image and event data: \textbf{Series I} (image-first), \textbf{Series II} (event-first), and \textbf{Parallel} (simultaneous use).}
    \vspace{-1em}
    \label{fig:5}
\end{figure}

%% file: table/abla_role.tex
\begin{table}[h]
\caption{Ablation Study on Image, Event, and Module Roles (Fig. \ref{fig:3}).}
\label{Tab.4}
\vspace{-1em}
\begin{center}
\begin{tabular}{lccccc}
    \toprule
    \textbf{} & \textbf{PSNR} & \textbf{SSIM}\\
    \midrule
    Baseline  & 27.1199 & 0.8843\\
    W/o Image & 23.2375 & 0.8511\\
    W/o Event & 25.4549 & 0.8580\\
    W/o Image Guide & 26.7900 & 0.8844\\
    W/o Event Guide & 25.7602 & 0.8697\\
    W/o Channel Attention & 23.4301& 0.8260\\ 
    W/o DwConv & 26.0102& 0.8718\\ 
    \bottomrule
\end{tabular}
\end{center}
\vspace{-1em}
\end{table}

%% file: sec/5_conclusion.tex
\section{Conclusion}
In this paper, we present a new approach that combines event cameras with retinex theory to enhance low-light RGB images. Our ERetinex model marks a significant advancement, offering improved illumination estimation and image quality in challenging lighting conditions. The potential applications of this work are vast, particularly in surveillance, autonomous systems, and low-power devices where robust image processing is crucial. Future work will focus on refining our model for high-dynamic-range imaging and 
exploring its generalizability. \textbf{Acknowledgment.} This work was supported by the National Natural Science Foundation of China under Grant Nos.62327808 and U24A20252, and the Key Research and Development Program of Shaanxi Province of China under Grant Nos. 2024PT-ZCK-66, 2024CY2-GJHX-48, and 2024CY2-GJHX-41.

%% file: main.bbl
\begin{thebibliography}{10}
\providecommand{\url}[1]{#1}
\csname url@rmstyle\endcsname
\providecommand{\newblock}{\relax}
\providecommand{\bibinfo}[2]{#2}
\providecommand\BIBentrySTDinterwordspacing{\spaceskip=0pt\relax}
\providecommand\BIBentryALTinterwordstretchfactor{4}
\providecommand\BIBentryALTinterwordspacing{\spaceskip=\fontdimen2\font plus
\BIBentryALTinterwordstretchfactor\fontdimen3\font minus \fontdimen4\font\relax}
\providecommand\BIBforeignlanguage[2]{{%
\expandafter\ifx\csname l@#1\endcsname\relax
\typeout{** WARNING: IEEEtran.bst: No hyphenation pattern has been}%
\typeout{** loaded for the language `#1'. Using the pattern for}%
\typeout{** the default language instead.}%
\else
\language=\csname l@#1\endcsname
\fi
#2}}

\bibitem{land1971lightness}
E.~H. Land and J.~J. McCann, ``Lightness and retinex theory,'' \emph{Josa}, vol.~61, no.~1, pp. 1--11, 1971.

\bibitem{Zhu_Zhang_Shen_Ma_Zhao_Zhou_2020}
\BIBentryALTinterwordspacing
A.~Zhu, L.~Zhang, Y.~Shen, Y.~Ma, S.~Zhao, and Y.~Zhou, ``\BIBforeignlanguage{en-US}{Zero-shot restoration of underexposed images via robust retinex decomposition},'' in \emph{\BIBforeignlanguage{en-US}{2020 IEEE International Conference on Multimedia and Expo (ICME)}}, Jul 2020. [Online]. Available: \url{http://dx.doi.org/10.1109/icme46284.2020.9102962}
\BIBentrySTDinterwordspacing

\bibitem{Wang_Zhang_Fu_Shen_Zheng_Jia_2019}
\BIBentryALTinterwordspacing
R.~Wang, Q.~Zhang, C.-W. Fu, X.~Shen, W.-S. Zheng, and J.~Jia, ``\BIBforeignlanguage{en-US}{Underexposed photo enhancement using deep illumination estimation},'' in \emph{\BIBforeignlanguage{en-US}{2019 IEEE/CVF Conference on Computer Vision and Pattern Recognition (CVPR)}}, Jun 2019. [Online]. Available: \url{http://dx.doi.org/10.1109/cvpr.2019.00701}
\BIBentrySTDinterwordspacing

\bibitem{Wei_Wang_Yang_Liu_2018}
C.~Wei, W.~Wang, W.~Yang, and J.~Liu, ``\BIBforeignlanguage{en-US}{Deep retinex decomposition for low-light enhancement},'' \emph{\BIBforeignlanguage{en-US}{arXiv: Computer Vision and Pattern Recognition,arXiv: Computer Vision and Pattern Recognition}}, Aug 2018.

\bibitem{Cai_2023_ICCV}
Y.~Cai, H.~Bian, J.~Lin, H.~Wang, R.~Timofte, and Y.~Zhang, ``Retinexformer: One-stage retinex-based transformer for low-light image enhancement,'' in \emph{Proceedings of the IEEE/CVF International Conference on Computer Vision (ICCV)}, October 2023, pp. 12\,504--12\,513.

\bibitem{Wu_2023_CVPR}
Y.~Wu, C.~Pan, G.~Wang, Y.~Yang, J.~Wei, C.~Li, and H.~T. Shen, ``Learning semantic-aware knowledge guidance for low-light image enhancement,'' in \emph{Proceedings of the IEEE/CVF Conference on Computer Vision and Pattern Recognition (CVPR)}, June 2023, pp. 1662--1671.

\bibitem{Fu_2023_CVPR}
H.~Fu, W.~Zheng, X.~Meng, X.~Wang, C.~Wang, and H.~Ma, ``You do not need additional priors or regularizers in retinex-based low-light image enhancement,'' in \emph{Proceedings of the IEEE/CVF Conference on Computer Vision and Pattern Recognition (CVPR)}, June 2023, pp. 18\,125--18\,134.

\bibitem{Jiang_Wang_Li_Zhang_Zhao_Gao}
Y.~Jiang, Y.~Wang, S.~Li, Y.~Zhang, M.~Zhao, and Y.~Gao, ``Event-based low-illumination image enhancement,'' \emph{IEEE Transactions on Multimedia}, pp. 1--12, 2023.

\bibitem{Lu_2023_CVPR}
Y.~Lu, Z.~Wang, M.~Liu, H.~Wang, and L.~Wang, ``Learning spatial-temporal implicit neural representations for event-guided video super-resolution,'' in \emph{Proceedings of the IEEE/CVF Conference on Computer Vision and Pattern Recognition (CVPR)}, June 2023, pp. 1557--1567.

\bibitem{Luo_2023_ICCV}
X.~Luo, K.~Luo, A.~Luo, Z.~Wang, P.~Tan, and S.~Liu, ``Learning optical flow from event camera with rendered dataset,'' in \emph{Proceedings of the IEEE/CVF International Conference on Computer Vision (ICCV)}, October 2023, pp. 9847--9857.

\bibitem{Zhu_2023_ICCV}
Z.~Zhu, J.~Hou, and D.~O. Wu, ``Cross-modal orthogonal high-rank augmentation for rgb-event transformer-trackers,'' in \emph{Proceedings of the IEEE/CVF International Conference on Computer Vision (ICCV)}, October 2023, pp. 22\,045--22\,055.

\bibitem{Sun_2023_CVPR}
L.~Sun, C.~Sakaridis, J.~Liang, P.~Sun, J.~Cao, K.~Zhang, Q.~Jiang, K.~Wang, and L.~Van~Gool, ``Event-based frame interpolation with ad-hoc deblurring,'' in \emph{Proceedings of the IEEE/CVF Conference on Computer Vision and Pattern Recognition (CVPR)}, June 2023, pp. 18\,043--18\,052.

\bibitem{Guo_Li_Ling_2017}
\BIBentryALTinterwordspacing
X.~Guo, Y.~Li, and H.~Ling, ``\BIBforeignlanguage{en-US}{Lime: Low-light image enhancement via illumination map estimation},'' \emph{\BIBforeignlanguage{en-US}{IEEE Transactions on Image Processing}}, p. 982–993, Feb 2017. [Online]. Available: \url{http://dx.doi.org/10.1109/tip.2016.2639450}
\BIBentrySTDinterwordspacing

\bibitem{Lee_Shih_Lien_Han_2013}
\BIBentryALTinterwordspacing
C.-H. Lee, J.-L. Shih, C.-C. Lien, and C.-C. Han, ``\BIBforeignlanguage{en-US}{Adaptive multiscale retinex for image contrast enhancement},'' in \emph{\BIBforeignlanguage{en-US}{2013 International Conference on Signal-Image Technology \&amp; Internet-Based Systems}}, Dec 2013. [Online]. Available: \url{http://dx.doi.org/10.1109/sitis.2013.19}
\BIBentrySTDinterwordspacing

\bibitem{Ren_Li_Cheng_Liu_2018}
X.~Ren, M.~Li, W.-H. Cheng, and J.~Liu, ``\BIBforeignlanguage{en-US}{Joint enhancement and denoising method via sequential decomposition},'' \emph{\BIBforeignlanguage{en-US}{Cornell University - arXiv,Cornell University - arXiv}}, Apr 2018.

\bibitem{Wang_Zheng_Hu_Li_2013}
\BIBentryALTinterwordspacing
S.~Wang, J.~Zheng, H.-M. Hu, and B.~Li, ``\BIBforeignlanguage{en-US}{Naturalness preserved enhancement algorithm for non-uniform illumination images},'' \emph{\BIBforeignlanguage{en-US}{IEEE Transactions on Image Processing}}, p. 3538–3548, Sep 2013. [Online]. Available: \url{http://dx.doi.org/10.1109/tip.2013.2261309}
\BIBentrySTDinterwordspacing

\bibitem{Jobson_Rahman_Woodell_1997}
\BIBentryALTinterwordspacing
D.~Jobson, Z.~Rahman, and G.~Woodell, ``\BIBforeignlanguage{en-US}{Properties and performance of a center/surround retinex},'' \emph{\BIBforeignlanguage{en-US}{IEEE Transactions on Image Processing}}, p. 451–462, Mar 1997. [Online]. Available: \url{http://dx.doi.org/10.1109/83.557356}
\BIBentrySTDinterwordspacing

\bibitem{Li_Liu_Yang_Sun_Guo_2018}
\BIBentryALTinterwordspacing
M.~Li, J.~Liu, W.~Yang, X.~Sun, and Z.~Guo, ``\BIBforeignlanguage{en-US}{Structure-revealing low-light image enhancement via robust retinex model},'' \emph{\BIBforeignlanguage{en-US}{IEEE Transactions on Image Processing}}, p. 2828–2841, Jun 2018. [Online]. Available: \url{http://dx.doi.org/10.1109/tip.2018.2810539}
\BIBentrySTDinterwordspacing

\bibitem{Zhang_Zhang_Guo_2019}
Y.~Zhang, J.~Zhang, and X.~Guo, ``\BIBforeignlanguage{en-US}{Kindling the darkness: A practical low-light image enhancer},'' \emph{\BIBforeignlanguage{en-US}{arXiv: Computer Vision and Pattern Recognition,arXiv: Computer Vision and Pattern Recognition}}, May 2019.

\bibitem{Yi_2023_ICCV}
X.~Yi, H.~Xu, H.~Zhang, L.~Tang, and J.~Ma, ``Diff-retinex: Rethinking low-light image enhancement with a generative diffusion model,'' in \emph{Proceedings of the IEEE/CVF International Conference on Computer Vision (ICCV)}, October 2023, pp. 12\,302--12\,311.

\bibitem{Huang_2023_CVPR}
X.~Huang, Y.~Zhang, and Z.~Xiong, ``Progressive spatio-temporal alignment for efficient event-based motion estimation,'' in \emph{Proceedings of the IEEE/CVF Conference on Computer Vision and Pattern Recognition (CVPR)}, June 2023, pp. 1537--1546.

\bibitem{zhang2020learning}
S.~Zhang, Y.~Zhang, Z.~Jiang, D.~Zou, J.~Ren, and B.~Zhou, ``Learning to see in the dark with events,'' in \emph{Computer Vision--ECCV 2020: 16th European Conference, Glasgow, UK, August 23--28, 2020, Proceedings, Part XVIII 16}.\hskip 1em plus 0.5em minus 0.4em\relax Springer, 2020, pp. 666--682.

\bibitem{Zhou_Teng_Han_Xu_Shi_2021}
\BIBentryALTinterwordspacing
C.~Zhou, M.~Teng, J.~Han, C.~Xu, and B.~Shi, ``\BIBforeignlanguage{en-US}{Delieve-net: Deblurring low-light images with light streaks and local events},'' in \emph{\BIBforeignlanguage{en-US}{2021 IEEE/CVF International Conference on Computer Vision Workshops (ICCVW)}}, Oct 2021. [Online]. Available: \url{http://dx.doi.org/10.1109/iccvw54120.2021.00135}
\BIBentrySTDinterwordspacing

\bibitem{liu2023low}
L.~Liu, J.~An, J.~Liu, S.~Yuan, X.~Chen, W.~Zhou, H.~Li, Y.~F. Wang, and Q.~Tian, ``Low-light video enhancement with synthetic event guidance,'' in \emph{Proceedings of the AAAI Conference on Artificial Intelligence}, vol.~37, no.~2, 2023, pp. 1692--1700.

\bibitem{liang2023coherent}
J.~Liang, Y.~Yang, B.~Li, P.~Duan, Y.~Xu, and B.~Shi, ``Coherent event guided low-light video enhancement,'' in \emph{Proceedings of the IEEE/CVF International Conference on Computer Vision}, 2023, pp. 10\,615--10\,625.

\bibitem{rebecq2019high}
H.~Rebecq, R.~Ranftl, V.~Koltun, and D.~Scaramuzza, ``High speed and high dynamic range video with an event camera,'' \emph{IEEE transactions on pattern analysis and machine intelligence}, vol.~43, no.~6, pp. 1964--1980, 2019.

\bibitem{cadena2021spade}
P.~R.~G. Cadena, Y.~Qian, C.~Wang, and M.~Yang, ``Spade-e2vid: Spatially-adaptive denormalization for event-based video reconstruction,'' \emph{IEEE Transactions on Image Processing}, vol.~30, pp. 2488--2500, 2021.

\bibitem{maharjan2019improving}
P.~Maharjan, L.~Li, Z.~Li, N.~Xu, C.~Ma, and Y.~Li, ``Improving extreme low-light image denoising via residual learning,'' in \emph{2019 IEEE international conference on multimedia and expo (ICME)}.\hskip 1em plus 0.5em minus 0.4em\relax IEEE, 2019, pp. 916--921.

\bibitem{Chen_2018_CVPR}
C.~Chen, Q.~Chen, J.~Xu, and V.~Koltun, ``Learning to see in the dark,'' in \emph{Proceedings of the IEEE Conference on Computer Vision and Pattern Recognition (CVPR)}, June 2018.

\bibitem{gu2019self}
S.~Gu, Y.~Li, L.~V. Gool, and R.~Timofte, ``Self-guided network for fast image denoising,'' in \emph{Proceedings of the IEEE/CVF International Conference on Computer Vision}, 2019, pp. 2511--2520.

\bibitem{lamba2021restoring}
M.~Lamba and K.~Mitra, ``Restoring extremely dark images in real time,'' in \emph{Proceedings of the IEEE/CVF Conference on Computer Vision and Pattern Recognition}, 2021, pp. 3487--3497.

\bibitem{xu2022snr}
X.~Xu, R.~Wang, C.-W. Fu, and J.~Jia, ``Snr-aware low-light image enhancement,'' in \emph{Proceedings of the IEEE/CVF conference on computer vision and pattern recognition}, 2022, pp. 17\,714--17\,724.

\bibitem{Chen_2023_CVPR}
X.~Chen, H.~Li, M.~Li, and J.~Pan, ``Learning a sparse transformer network for effective image deraining,'' in \emph{Proceedings of the IEEE/CVF Conference on Computer Vision and Pattern Recognition (CVPR)}, June 2023, pp. 5896--5905.

\bibitem{wang2023ultra}
T.~Wang, K.~Zhang, T.~Shen, W.~Luo, B.~Stenger, and T.~Lu, ``Ultra-high-definition low-light image enhancement: A benchmark and transformer-based method,'' in \emph{Proceedings of the AAAI Conference on Artificial Intelligence}, vol.~37, no.~3, 2023, pp. 2654--2662.

\bibitem{Fu_Zeng_Huang_Zhang_Ding_2016}
\BIBentryALTinterwordspacing
X.~Fu, D.~Zeng, Y.~Huang, X.-P. Zhang, and X.~Ding, ``\BIBforeignlanguage{en-US}{A weighted variational model for simultaneous reflectance and illumination estimation},'' in \emph{\BIBforeignlanguage{en-US}{2016 IEEE Conference on Computer Vision and Pattern Recognition (CVPR)}}, Jun 2016. [Online]. Available: \url{http://dx.doi.org/10.1109/cvpr.2016.304}
\BIBentrySTDinterwordspacing

\bibitem{Kingma_Ba_2014}
D.~Kingma and J.~Ba, ``\BIBforeignlanguage{en-US}{Adam: A method for stochastic optimization},'' \emph{\BIBforeignlanguage{en-US}{arXiv: Learning,arXiv: Learning}}, Dec 2014.

\bibitem{Loshchilov_Hutter_2016}
I.~Loshchilov and F.~Hutter, ``\BIBforeignlanguage{en-US}{Sgdr: Stochastic gradient descent with warm restarts},'' \emph{\BIBforeignlanguage{en-US}{International Conference on Learning Representations,International Conference on Learning Representations}}, Aug 2016.

\end{thebibliography}
